\pdfoutput=1

\documentclass[11pt]{article}

\usepackage[preprint]{acl}

\usepackage{times}
\usepackage{latexsym}

\usepackage[T1]{fontenc}

\usepackage[utf8]{inputenc}

\usepackage{microtype}

\usepackage[scaled=0.9]{beramono}
\usepackage{soul}

\usepackage{graphicx}

\usepackage{enumitem}
\usepackage{booktabs}
\usepackage{subcaption}  
\usepackage{multirow}

\usepackage{commath}
\usepackage[normalem]{ulem}

\title{Cross-Lingual Representation Alignment Through Contrastive Image-Caption Tuning}

\author{
 \textbf{Nathaniel Krasner},
 \textbf{Nicholas Lanuzo},
 \textbf{Antonios Anastasopoulos}
\\
 Department of Computer Science, George Mason University; Fairfax, VA.
\\
 \small{
   \textbf{Correspondence:} \href{nkrasner@gmu.edu}{nkrasner@gmu.edu}
 }
}

\begin{document}
\maketitle
\begin{abstract}
Multilingual alignment of sentence representations has mostly required bitexts to bridge the gap between languages. We investigate whether visual information can bridge this gap instead. Image caption datasets are very easy to create without requiring multilingual expertise, so this offers a more efficient alternative for low-resource languages. We find that multilingual image-caption alignment can implicitly align the text representations between languages, languages unseen by the encoder in pretraining can be incorporated into this alignment post-hoc, and these aligned representations are usable for cross-lingual Natural Language Understanding (NLU) and bitext retrieval. \footnote{Data and code will be publicly released at \url{https://github.com/nkrasner/cl-clip-align}.}
\end{abstract}

\section{Introduction}

Encoder language models are still a very popular and widely used method for extracting semantic information from text to be used downstream for natural language understanding (NLU) tasks. In general, an encoder language model (LM) is pretrained on a large corpus using self-supervision and then a smaller component is fine-tuned on annotated data using the representations produced by the pretrained LM. For widely spoken data-rich languages, this is no problem and the existence of task-specific, annotated data is a given~\cite{joshi-etal-2020-state,blasi-etal-2022-systematic}. For low-resource languages this is rarely the case, as collecting data for each task in such languages is expensive and time consuming. Thus, cross-lingual knowledge transfer is a more practical direction for low-resource languages than further data collection. %

While multilingual encoder-only LMs have recently fallen out of favor compared to their large decoder-only counterparts, speakers from under-served communities express a need for proper language understanding in their languages, more than any need for technologies that generate language~\cite{blaschke-etal-2024-dialect}. %

The internal representations of encoder models trained on multilingual data tend to be disjoint, so the representation of a sentence in language A may not be similar to the representation of its translation in language B. Most likely, this is the result of pretraining data imbalance and domain mismatch across the languages included in their pretraining. If these internal representations were aligned such that representations of translations \textit{were} similar, cross-lingual transfer for NLU tasks should be much easier to achieve, as %
\citet{hu-etal-2021-explicit} showed. 
This cross-lingual transfer of task knowledge can greatly benefit speakers of low-resource languages by giving them access to NLP tools without the difficulty of annotating task-specific data in their language. As an additional benefit, these aligned representations can be used to mine bitexts from large scraped corpora to build parallel translation datasets \cite{nllbteam2022languageleftbehindscaling}.

In this work, we explore whether one could encourage multilingual representation alignment \textit{without any parallel data}, but relying instead on images as a grounding, shared modality across languages. 
This is a worthwhile direction to pursue for two reasons. First, parallel text curation through expert translation is time-consuming and expensive. In contrast, it is easy for any language speaker to describe an image to produce a caption~\cite{madaan2020practicalcomparabledatacollection}. %
Second, language documentation efforts often produce media accompanied with monolingual audio or text in the language of interest. %
Developing techniques that would leverage such materials could enable the creation of technologies for these otherwise under-served languages.

To summarize, we (1) show that a multilingual text-image contrastive learning setup can produce aligned representations; (2) focus specifically on Quechua, as an example of a language unseen during pretraining that may benefit from such approaches; and (3) show that this method does not degrade representation quality in other languages.

\section{Related Work}
Previous endeavors in multilingual alignment in the absence of parallel-text supervision have predominantly concentrated on the alignment of static-word embeddings through adversarial techniques \cite{zhang-etal-2017-adversarial,chen-cardie-2018-unsupervised}. Approaches that extend multilingual alignment to sentence-level representations have generally necessitated a bitext signal \cite{feng-etal-2022-language,escolano-etal-2021-multilingual,artetxe-etal-laser}, with limited exceptions employing adversarial methodologies \cite{aghajanyan-etal-2019-towards,tien-steinert-threlkeld-2022-bilingual}. Even though multilingual alignment may extend to languages not encountered during fine-tuning \cite{tien-steinert-threlkeld-2022-bilingual}, we hypothesize that a more direct fine-tuning strategy using some pivot (even if not textual) could potentially produce superior alignment for languages with limited bitext resources.

Contrastive methods have been used for text-text \cite{feng-etal-2022-language} encoder alignment as well text-image encoder alignment in both the monolingual \cite{pmlr-v139-radford21a} and multilingual \cite{10.1145/3581783.3611992,bianchi2021contrastivelanguageimagepretrainingitalian} setting. One such text-image alignment work introduces an image representation into the input sequence of NLU tasks leading to improved cross-lingual transfer \cite{10.1145/3581783.3611992}. This offers additional support to our hypothesis that visual information can act as a semantic bridge between languages.

\section{Method, Experiments, and Results}
\label{sec:method}

Our approach strings together a text encoder with a vision encoder. These two produce representations for each modality input, which are then used in a contrastive learning setup. In  particular, given pairs of image representations $E_i$ and caption representations $E_c$ we use the following, simple contrastive loss function:\\[-2em]
\begin{align*}
S &= E_c \cdot E_i^\top * t\\
L(E_i,E_c) &= \text{CrossEntropy}(S, I),
\end{align*}
where $I$ is the identity matrix and $t$ is a learned temperature parameter.

This is similar to what CLIP \cite{pmlr-v139-radford21a} used for text-image alignment and LaBSE \cite{feng-etal-2022-language} for text-text alignment.

\subsection{Experimental Setup}
\label{subsec:setup}

\paragraph{Datasets}
We work with the MS-COCO dataset~\cite{lin2015microsoftcococommonobjects}, which provides 118k English Image-Caption pairs.
Using Google Translate, we translate the English captions into Spanish, Japanese, Hindi, and Quechua. %
From this 5-way parallel image caption dataset, we now derive 4 datasets for various experiments:
\begin{enumerate}[noitemsep,nolistsep]
    \item \texttt{Eng-only}: The plain MS-COCO dataset without translations to other languages.
    \item \texttt{Eng-Pivot}: The English captions from MS-COCO paired with one translation per sample from a rotation of Spanish, Japanese, Hindi.
    \item \texttt{Multilingual}: The MS-COCO dataset but each caption is from a rotation of English, Spanish, Japanese, and Hindi with only one language paired with each image.
    \item \texttt{Multilingual+Quechua}: The same as the Multilingual dataset but with Quechua added into the rotation of languages.
\end{enumerate}
While the other datasets are designed for use with text-image alignment, the \texttt{Eng-Pivot} dataset is used for text-text alignment to create a model similar to LaBSE \cite{feng-etal-2022-language} with a comparable data size to our other models.

\paragraph{Training}
We fine-tune an XLM-Roberta-Large (XLM-R) \cite{conneau2020unsupervisedcrosslingualrepresentationlearning} text encoder and a VIT-Base-patch16-224-in21k \cite{dosovitskiy2021imageworth16x16words} image encoder. 

The token-level representations are mean pooled to create a sentence-level representation. 
Since the hidden dimensions of these encoders do not match, we add a linear layer to their outputs to adapt them to a matching dimensionality of 512. Following existing approaches to text-image alignment under these circumstances \cite{bianchi2021contrastivelanguageimagepretrainingitalian}, we allow these linear layers to warm up for a certain number of steps before fine-tuning the encoders themselves. In our case, we chose to "thaw" the encoders halfway through the first epoch since the learning curves had flattened out by that point.

\subsection{Experiment 1: Does multilingual text-image alignment lead to text-text alignment?}

We hypothesize that text-image alignment involving multiple languages will implicitly align text representations between languages.

With the exception of the Eng-Pivot encoder (which is trained on bitext alignment), our encoders are only fine-tuned to align the text representations to the image representations, but we evaluate them on their alignment between text representations.
Specifically, we use the Flores-200 dataset \cite{nllbteam2022languageleftbehindscaling}, which contains 200-way parallel sentences including all our test languages.
We perform a formal analysis using the task of bitext retrieval \cite{heffernan-etal-2022-bitext, duquenne2023sonarsentencelevelmultimodallanguageagnostic} as well as a visual analysis via t-SNE.

We compare against a baseline of the off-the-shelf XLM-R encoder, as well as one fine-tuned on our text-image pretraining using only the English-only dataset, and another trained directly on contrastive text-text alignment with an English pivot similarly to LaBSE \cite{feng-etal-2022-language}.

For each sentence in each language, we search the English sentences in Flores-200 for the minimum cosine distance to find a match (translation). If the translation selected is the true translation, we count that sentence as correct. We calculate the retrieval accuracy over each language and then aggregate using the mean over all languages to produce a final score.
Since the language encoder has not seen all of these languages in pretraining, we report the retrieval accuracy over the disjoint subsets of languages on which it was trained (or not).

\begin{table}
    \centering
    \resizebox{0.48\textwidth}{!}{
    \begin{tabular}{r|cccc}
    \toprule
        & All & in XLM-R & not in XLM-R & \\
        Encoder & (203 langs) & (92 langs) & (111 langs) & Quechua \\
        \midrule
        XLM-R & 0.5 & 0.6 & 0.4 & 0.5 \\
        Eng-Only & 18.3 & 27.5 & 10.7 & 7.2 \\
        Eng-Pivot & \textbf{62.2} & \textbf{92.6} & \textbf{37.1} & 13.1 \\
        Multilingual & 55.7 & 82.2 & 33.7 & 18.0 \\
        + Quechua & 50.4 & 76.6 & 28.6 & \textbf{29.2} \\
        \bottomrule
    \end{tabular}
    }
    \vspace{-.5em}
    \caption{Bitext retrieval accuracy on All of flores-200, on the subset of languages in/not in XLM-R's pretraining, and just on Quechua.}
    \label{tab:bitext-acc}
    \vspace{-1.5em}
\end{table}

While not quite matching the text-text aligned encoder (Eng-Pivot in table \ref{tab:bitext-acc}), the multilingual text-image aligned encoder (Multilingual in table \ref{tab:bitext-acc}) is still very capable in the bi-text retrieval task. 
The English-Only text-image alignment improves on the abysmal results of the plain XLM-R model, but does not compare with the multilingual alignment. This is likely because the pretraining of XLM-R does not scale to sentence level tasks well \cite{reimers2019sentencebertsentenceembeddingsusing}. The text-image alignment, on its own, may expand the existing knowledge of XLM-R to the sentence level.

\begin{figure*}
    \centering
    \includegraphics[width=.9\textwidth]{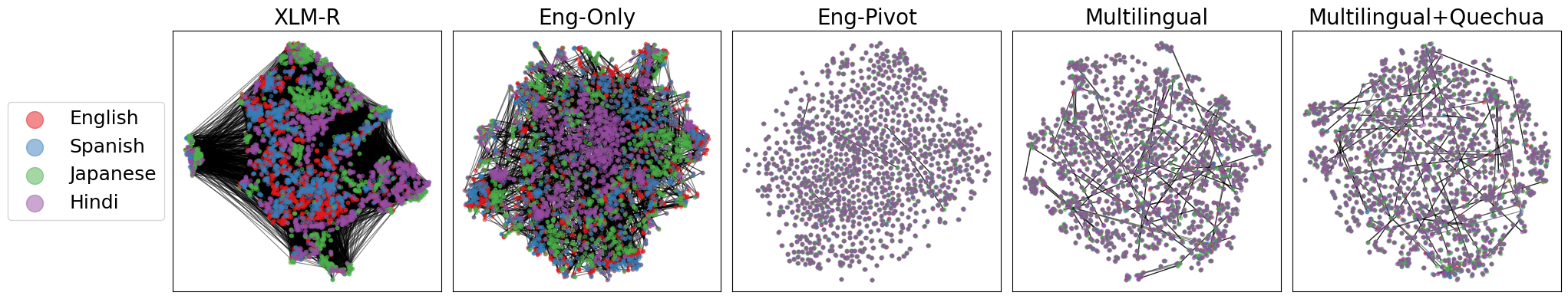}
    \vspace{-1em}
    \caption{t-SNE embeddings for the outputs of each encoder over flores-200 sentences. Translations are shown as cliques with lines connecting them. Visible lines, such as those in the two leftmost panels, indicate that representations of translated sentences are far from each other, ie., \textit{poor} alignment. While not as clear as parallel text alignment (Eng-Pivot), multilingual image-text alignment (two rightmost panels) shows promising results.} %
    \label{fig:tsne-plots}
    \vspace{-1em}
\end{figure*}

To further visualize the multilingual alignment of our encoders we generate sentence-level representations for all sentences in the Flores-200 dataset and use $t$-SNE to project them down to~2 dimensions while preserving relative distances. We plot these embeddings in Figure~\ref{fig:tsne-plots} for the~4 fine-tuning languages with lines connecting parallel cliques of translated sentences. This way we can visualize whether an encoder produces language-specific clusters or whether certain sentences are encoded far from their translations.

Figure \ref{fig:tsne-plots} shows that the original XLM-R representations are not aligned at all. Tuning only on the English image captions, although better than the untuned model, the languages still form distinct clusters. Our Multilingual approach falls just short of the text-text aligned model in terms of the number of misaligned translations and adding Quechua into the mix does not make it that much worse. Interestingly, the text-image aligned models have tighter clusters indicating that the image alignment may have drawn connections between sentences that were not there previously. %

\subsection{Experiment 2: Can a language unseen in the encoder's pretraining be added using only image caption tuning?}

Here we turn to investigating the possibility of using only caption-text data to obtain good representations for a language \textit{unseen} during pretraining, without any other parallel text data.
This approximates a real setting where we have access to a newspaper or similar dataset with image captions in a low-resource language and we want to add it to the aligned language encoder for use in downstream tasks in a zero-shot cross-lingual transfer setting~\cite{madaan2020practicalcomparabledatacollection}.

We find that languages not included in the pretraining or fine-tuning still benefit from some alignment. But as one would expect, not to the same degree as those which have been already included in the model's training data. %

We retrained the encoder from Experiment 1, but now with a dataset that also mixes in Quechua captions. Indigenous Latin American languages, including Quechua, are not included in the pretraining data of XLM-R. Quechua is also typologically distinct from all other pretraining languages. %

We calculate the retrieval accuracy on Flores-200 from Quechua to English as well as the overall X$\rightarrow$English accuracy to determine how well Quechua has been integrated into the encoder and aligned with other languages. 

When Quechua is added to the image-caption dataset, the overall performance goes down, but the performance on Quechua is greatly improved (cf last two rows of Table~\ref{tab:bitext-acc}) from 18\% to 29.2\%. Importantly, the average accuracy for all other languages remains largely unaffected -- we attribute the small drop in performance to the fact that we reduced the data in the other four languages to ensure experimental data-size comparability; in practice, this is not a requirement in the real world.

\subsection{Experiment 3: Are the downstream qualities of the representations preserved and is cross-lingual transfer possible?}

Here we go beyond intrinsic evaluation to test our embeddings for a downstream task:  natural language inference (NLI). Since images and text contain different types of semantic information, we want to ensure that aligning a text encoder to an image encoder does not overwrite the features which are useful for downstream NLU tasks.

We train simple feed-forward NLI models on frozen representations from each of the models in the previous experiments using the combined MultiNLI training and dev sets. 

We train using the MultiNLI train and dev datasets which only contain English samples. Any samples marked by the authors as lacking agreement were discarded. For evaluation of downstream NLI quality, we use the XNLI test portion to measure both English NLI and cross-lingual transfer performance.

For each encoder, we train identical NLI models with input features ($\oplus$ stands for concatenation):\\[-2em]
\begin{align*}
x_i = & \ e(p_i) \ \oplus \ e(h_i) \oplus
|e(p_i) - e(h_i)| \\ & \oplus e(p_i) * e(h_i)
\end{align*}
where $e$ is the encoder and $p_i$ and $h_i$ are a premise and hypothesis respectively \cite{conneau-etal-2017-supervised}.

The NLI models are a simple feed-forward architecture with 2 hidden layers and a hidden size of 2048. They are trained using the Adam optimizer and a learning rate of $2*10^{-5}$ for 100 epochs with early stopping.

The results in Table \ref{tab:nli_acc} show that the alignment of the text encoder with the space of the image encoder does not damage the quality of the text representations for downstream use, but actually improves them. 
Comparing the Multilingual image aligned model before and after adding Quechua, downstream performance is somewhat uncoupled from bitext retrieval performance. The addition of Quechua matched or exceeded the performance without it across nearly all languages, suggesting that NLI performance benefits from increased language coverage regardless of individual language data size. English represents $\frac{1}{4}$ of the Multilingual dataset and $\frac{1}{5}$ after adding Quechua, but the addition of Quechua increased the NLI score on English! Additionally, fine-tuning the encoder on the Eng-Only dataset only made a minimal improvement to the XLM-R performance even though it saw the largest portion of English data.

\begin{table}[t]
    \centering
    \footnotesize
    \begin{tabular}{@{}rr@{ }r@{ }r@{ }r@{ }r@{ }r@{ }r@{ }r@{ }r@{ }r@{ }r@{ }r|r@{}}
    \toprule
        Encoder & ar  & de & el & \underline{en} & \underline{es} &  \underline{hi} & ru & sw & th & tr & ur & zh & Avg  \\
        \midrule
        XLM-R & 45  & 43 & 43 & 50 & 44  & 44 & 44 & 37 & 42 & 43 & 42 & 44 & 43.8 \\
        Eng-Only & 47 & 49 & 48 & \underline{53} & 50 & 46 & 50 & 42 & 47 & 47 & 45  & 48 & 48\\
        Eng-Pivot & 61 & 64 & 63 & \underline{67} & \underline{65} & \underline{60} & 62 & 52 & 61 & 61 & 58  & 62 & 61.8\\
        Multiling. & 51 & 52 & 52 & \underline{55} & \underline{52} &\underline{51} & 52 & 45 & 51 & 51 & 48  & 51 & 51.3\\
        + Quechua & 51 & 53 & 53 & \underline{56} & \underline{53} &  \underline{51} & 53 & 45 & 50 & 51 & 49 & 51 & 51.6\\
        \bottomrule
    \end{tabular}
    \caption{Rounded XNLI accuracy (on sample languages). Languages seen for alignment fine-tuning are underlined. NLI models are only trained on English data with frozen encoders; results in other languages require cross-lingual transfer.}
    \label{tab:nli_acc}
    \vspace{-1em}
\end{table}

\section{Conclusion}
The task of multilingual text-image contrastive alignment implicitly aligns text from multiple languages into the same space. This alignment carries over into unseen languages, and performance on a particular unseen language can be improved by collecting image-caption pairs in that language.

While this technique does not outperform SOTA methods, it performs remarkably well considering the non-reliance on parallel corpora. For low resource languages, this method could act as a boot-strapping step to scrape higher quality bitexts for use in further alignment.

\section{Limitations}
With the addition of Quechua to the training set, the drop in overall bitext retrieval performance could be due to the decrease in data for the other languages to accommodate the Quechua data. Whether this is the case is not captured by our experiment, but can be taken into account in a follow-up work.

\section{Acknowledgements}
This work was partially supported by resources provided by the Office of Research Computing at George Mason University (URL: https://orc.gmu.edu) and funded in part by grants from the National Science Foundation (Award Number 2018631).

\bibliography{anthology,custom}

\clearpage
\pagebreak

\appendix

\section{Complete XNLI Results}
\label{sec:completexnli}

Table~\ref{tab:nli_acc_complete} presents all our results in the XNLI test set.

\begin{table*}[t!]
    \centering
    \footnotesize
    \begin{tabular}{r|rrrrrrrrrrrrrrr|c}
    \toprule
        Encoder & ar & bg & de & el & \underline{en} & \underline{es} & fr & \underline{hi} & ru & sw & th & tr & ur & vi & zh & Avg  \\
        \midrule
        XLM-R & 45 & 43 & 43 & 43 & 50 & 44 & 47 & 44 & 44 & 37 & 42 & 43 & 42 & 46 & 44 & 43.8 \\
        Eng-Only & 47 & 49 & 49 & 48 & \underline{53} & 50 & 51 & 46 & 50 & 42 & 47 & 47 & 45 & 48 & 48 & 48\\
        Eng-Pivot & 61 & 63 & 64 & 63 & \underline{67} & \underline{65} & 65 & \underline{60} & 62 & 52 & 61 & 61 & 58 & 63 & 62 & 61.8\\
        Multilingual & 51 & 53 & 52 & 52 & \underline{55} & \underline{52} & 53 & \underline{51} & 52 & 45 & 51 & 51 & 48 & 52 & 51 & 51.3\\
        + Quechua & 51 & 53 & 53 & 53 & \underline{56} & \underline{53} & 53 & \underline{51} & 53 & 45 & 50 & 51 & 49 & 52 & 51 & 51.6\\
        \bottomrule
    \end{tabular}
    \caption{Rounded XNLI accuracy. Languages seen for alignment fine-tuning are underlined. NLI models are only trained on English data with frozen encoders; results in other languages require cross-lingual transfer.}
    \label{tab:nli_acc_complete}
\end{table*}
\end{document}